\documentclass[runningheads]{llncs}
\usepackage[T1]{fontenc}
\usepackage{graphicx}
\newcommand\TheText[1]{\colorbox[RGB]{208,226,236}{#1}}
\newcommand\RelationSet[1]{\colorbox{LemonChiffon2}{#1}}
\newcommand\EntitySet[1]{\colorbox{LavenderBlush2}{#1}}
\newcommand\EntityPairs[1]{\colorbox[RGB]{255,224,153}{#1}}
\newcommand\PredictRelationSet[1]{\colorbox{Honeydew2}{#1}}
\newcommand\Entity[1]{\colorbox[RGB]{220,208,255}{#1}}

\usepackage{graphicx}
\usepackage{amsmath}
\usepackage{array}
\usepackage{booktabs}
\usepackage{caption}
\usepackage{subcaption}
\usepackage{pgfplotstable}
\usepackage{multirow}
\usepackage{booktabs}

\usepackage[dvipsnames,svgnames,x11names]{xcolor}
\usepackage[most]{tcolorbox}
\newtcolorbox{mybox}[2][]{
	width=\columnwidth,
	colback = gray!8, 
	colframe = gray!8, 
	boxsep=0pt,left=10pt,right=10pt,top=0pt,bottom=0pt,
	fontupper=\linespread{0.9}\selectfont,
	title=#2,#1}

\begin{document}
\title{Relation as a Prior: A Novel Paradigm for LLM-based Document-level Relation Extraction}
\titlerunning{Relation as a Prior: A Novel Paradigm for LLM-based DocRE}

\author{
Qiankun Pi†\inst{1} \and
Yepeng Sun†\inst{2} \and
Jicang Lu\thanks{Corresponding author.}\inst{1} \and
Qinlong Fan\inst{1} \and
Ningbo Huang\inst{1} \and
Shiyu Wang\inst{1}
}

\authorrunning{Q. Pi et al.}

\institute{%
Information Engineering University, Zhengzhou, China\\
\email{lujicang@sina.com}\\
\email{\{piqiankun2000, fan\_qinlong, rylynn\_ab, share\_wind\}@163.com}
\and
Academy of Military Science, Beijing, China\\
\email{yepeng\_ah@163.com}
}

%
\maketitle              
\begin{abstract}
Large Language Models (LLMs) have demonstrated their remarkable capabilities in document understanding. However, recent research reveals that LLMs still exhibit performance gaps in Document-level Relation Extraction (DocRE) as requiring fine-grained comprehension. The commonly adopted "extract entities then predict relations" paradigm in LLM-based methods leads to these gaps due to two main reasons: (1) Numerous unrelated entity pairs introduce noise and interfere with the relation prediction for truly related entity pairs. (2) Although LLMs have identified semantic associations between entities, relation labels beyond the predefined set are still treated as prediction errors.
\par
To address these challenges, we propose a novel "\textbf{Rel}ation as a \textbf{Prior}" \textbf{(RelPrior)} paradigm for LLM-based DocRE. For challenge (1), RelPrior utilizes binary relation as a prior to extract and determine whether two entities are correlated, thereby filtering out irrelevant entity pairs and reducing prediction noise. For challenge (2), RelPrior utilizes predefined relation as a prior to match entities for triples extraction instead of directly predicting relation. Thus, it avoids misjudgment caused by strict predefined relation labeling. Extensive experiments on two benchmarks demonstrate that RelPrior achieves state-of-the-art performance, surpassing existing LLM-based methods.

\end{abstract}
\section{Introduction}
Relation Extraction (RE), a key task in natural language processing, aims to identify relationships between related entity pairs \cite{shi2019discovering,trisedya2019neural,yu2017improved}. In the application scenarios of real document contexts, a large number of valuable relational facts are usually expressed through multiple sentence. Current research is shifting its focus from sentence-level relation extraction to document-level relation extraction (DocRE), aiming to identify relationships between entities throughout the entire document \cite{peng2017cross,yao2019docred}.

\begin{figure}[t]
  \includegraphics[width=\columnwidth]{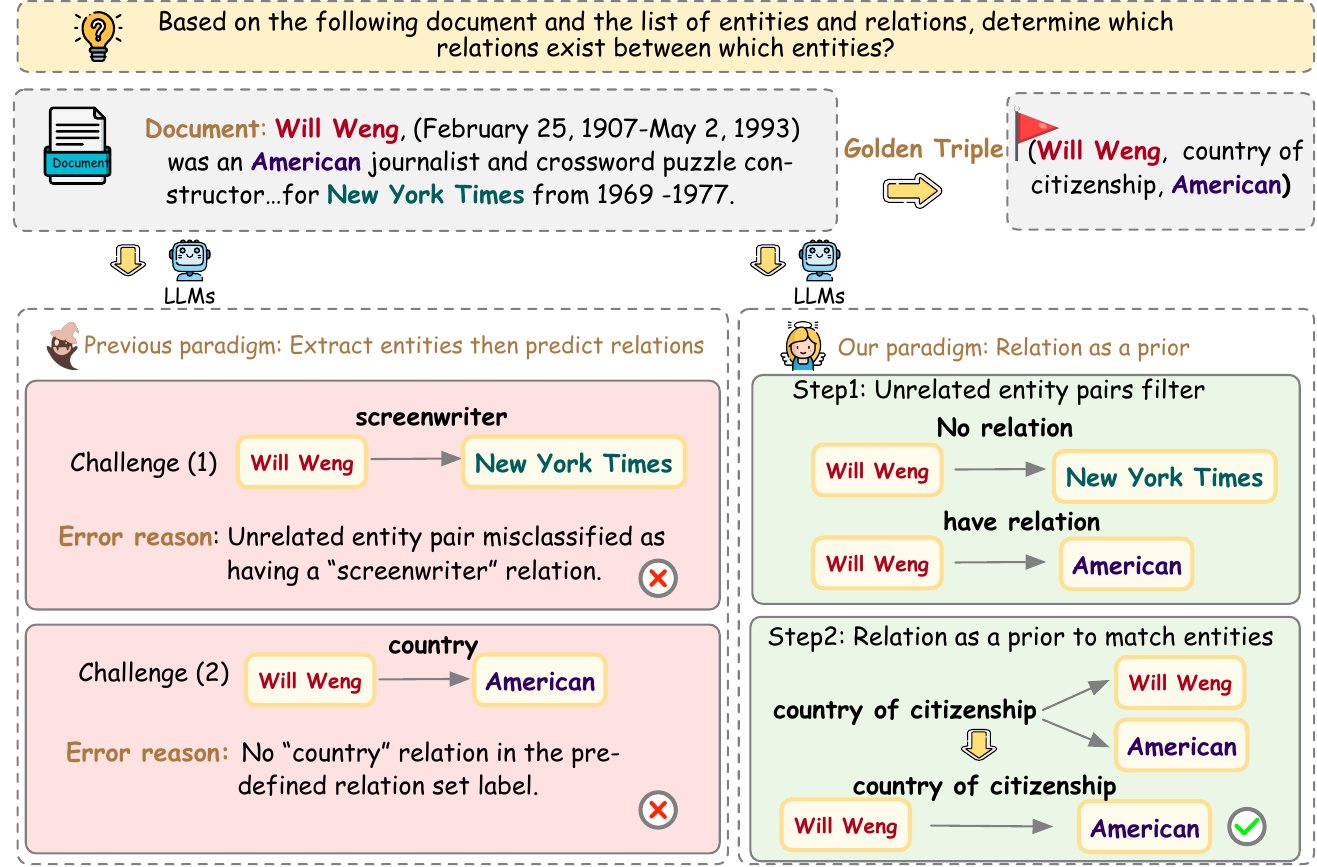}
  \caption{Comparison between existing paradigm "extract entities then predict relations" and our proposed paradigm "relation as a prior" for DocRE based on LLMs.}
  \label{Graph1}
\end{figure}

In the DocRE tasks, traditional methods based on pre-trained models primarily focus on modeling the relationships between entity pairs using text structures or semantic features. Existing studies can be broadly classified into two types of DocRE: Sequence Structure (SS-based) and Graph Structure (GS-based), depending on how the text is modeled.
SS-based methods model information such as document context, sentence structure, and entity semantic features, transforming relation prediction between entity pairs into the calculation of semantic similarity between vector representations, thereby enabling relation extraction, e.g., \cite{xie2022eider,xu2022document,zhou2021document}. GS-based methods use graph structure to represent the relationship between entities in the document, and identifies the relationship, e.g., \cite{ding2023clgr,liu2023document,zeng2020double}.

While large language models (LLMs) have significantly advanced the field of natural language processing \cite{Brown2020Language,qin2024large}, their performance in DocRE still lags behind that of traditional pre-trained models. This limitation has prompted researchers to explore more effective strategies for adapting LLMs to DocRE tasks. Specifically, \cite{li2023semi} introduced a hybrid method combining LLMs with Natural Language Inferencing (NLI) model to predict relation triples, addressing the issue of incorrect relation types in existing datasets. \cite{xue2024autore} proposed prioritizing relations before identifying corresponding triples, and their approach enhanced the performance of relation extraction. Additionally, \cite{tao2024graphical} proposed prompt designs supplemented with examples to optimize LLMs for DocRE tasks.

Although these LLM-based methods have shown improved performance, they all rely on the "Extract entities then predict relations" paradigm. Due to the disorderly nature of entities and the uncertainty of relations between them, there are two challenges when LLMs adopt this paradigm. (1) A significant proportion of entities in the document are unrelated and have no relationship with each other. As shown in Fig. \ref{Graph1} (challenge 1), "Will Weng" and "New York Times" are actually unrelated, but the LLMs ultimately predict that they are related as "screenwriter". (2) LLMs predict the relations between entity pairs, and their align semantically with the predefined relations. However, due to the strict relation labels, misjudgment may occur in the statistical results. As shown in Fig. \ref{Graph1} (challenge 2), LLMs predict the relation "country" with semantics consistent with the "country of citizenship" label, but it is classified as an error due to the strict labeling criteria.

To tackle these problems, we propose a novel "\textbf{Rel}ation as a \textbf{Prior}" \textbf{(RelPrior)} paradigm for LLM-based DocRE. Specifically, to address the challenge of unrelated entity pairs, we propose fine-tuning a binary classification LLM that determines whether there is a relation between entity pairs, prioritizing the transformation of relation extraction into a binary classification task to further filter out unrelated entity pairs. To address the challenge of relation misjudgment, we propose using relation as a prior to match entity pairs and determine which entities can serve as the head and tail entities for the given relation. This approach transforms the relation extraction task into a task of judging both entity pairs and relations, avoiding the arbitrary generation issues of LLM. 

In summary, our main contributions are as follows.

\begin{itemize}

\item[$\bullet$] We make the first attempt to explore a novel "Relation as a Prior" paradigm for LLM-based DocRE, which aims to enhance the ability of LLMs to extract relations and narrow the performance gap with traditional pre-trained models.
\item[$\bullet$] We propose transforming relation filtering into a binary classification task and relation extraction into a task of distinguishing between known relations and entity pairs, thereby avoiding the impact of generative hallucinations from LLMs on the final quality of the triplets.
\item[$\bullet$] Extensive experiments on two DocRE datasets show that RelPrior surpasses the existing DocRE methods LLM-based and achieves state-of-the-art performance.
\end{itemize}

\section{Related work}
\subsection{Relation Extraction with LLMs}
Relation Extraction based on LLMs leverages their powerful semantic understanding capabilities to better capture deep textual semantics, demonstrating excellent performance in contextual understanding for RE tasks. \cite{ma2023Chain,wadhwa2023revisiting} applied LLMs to Sentence-level RE with promising results. However, DocRE presents greater challenges than Sentence-level RE, featuring more complex relation types and more difficult relation prediction. Currently, LLMs' performance on DocRE tasks remains suboptimal, still lagging behind traditional methods in terms of effectiveness.

\cite{wang2024document-level} proposed a framework that employs retrieval-chain prompts to guide ChatGPT in progressively generating documents with complex semantic contexts and diverse relation triplets, effectively addressing the high cost of manual data annotation. To mitigate the generation of irrelevant content caused by LLMs hallucinations in such approaches, \cite{li2023semi} introduced a hybrid method combining LLMs with NLI models, thereby enhancing performance on DocRE tasks.

However, existing RE methods inadequately address the noise interference caused by irrelevant entity pairs in documents, often failing to effectively identify and filter them. This limitation highlights the need for a new DocRE method that can efficiently filter out irrelevant entity pairs to improve RE performance.

\subsection{Prompt Learning}
Prompt learning involves providing models with specific instructions or questions by optimizing prompt structures and formulations to guide desired outputs. \cite{tao2024graphical} conducted comprehensive research on prompt learning in LLMs, demonstrating that incorporating detailed prompt examples when using GPT-3.5 for reasoning significantly improves RE accuracy. Building on this foundation, \cite{Zhang2024PathofThoughtsEA} introduced a based thought-path-prompting approach that simulates human problem-solving strategies to better address complex RE tasks. As research progressed, scholars identified substantial limitations in relying solely on prompts. Consequently, \cite{wan2023gpt} proposed integrating reasoning logic into prompts, progressively enhancing LLMs' ability to discern entity relationships and improving their performance on RE tasks.

However, for complex DocRE tasks, although traditional prompt learning methods can predict semantic relationships that match the golden labels, they may still be judged as incorrect. To address this issue, it is necessary to develop a new prompt learning method to avoid label misjudgment.

\begin{figure*}[t]
  \centering
  \includegraphics[width=1.0\linewidth]{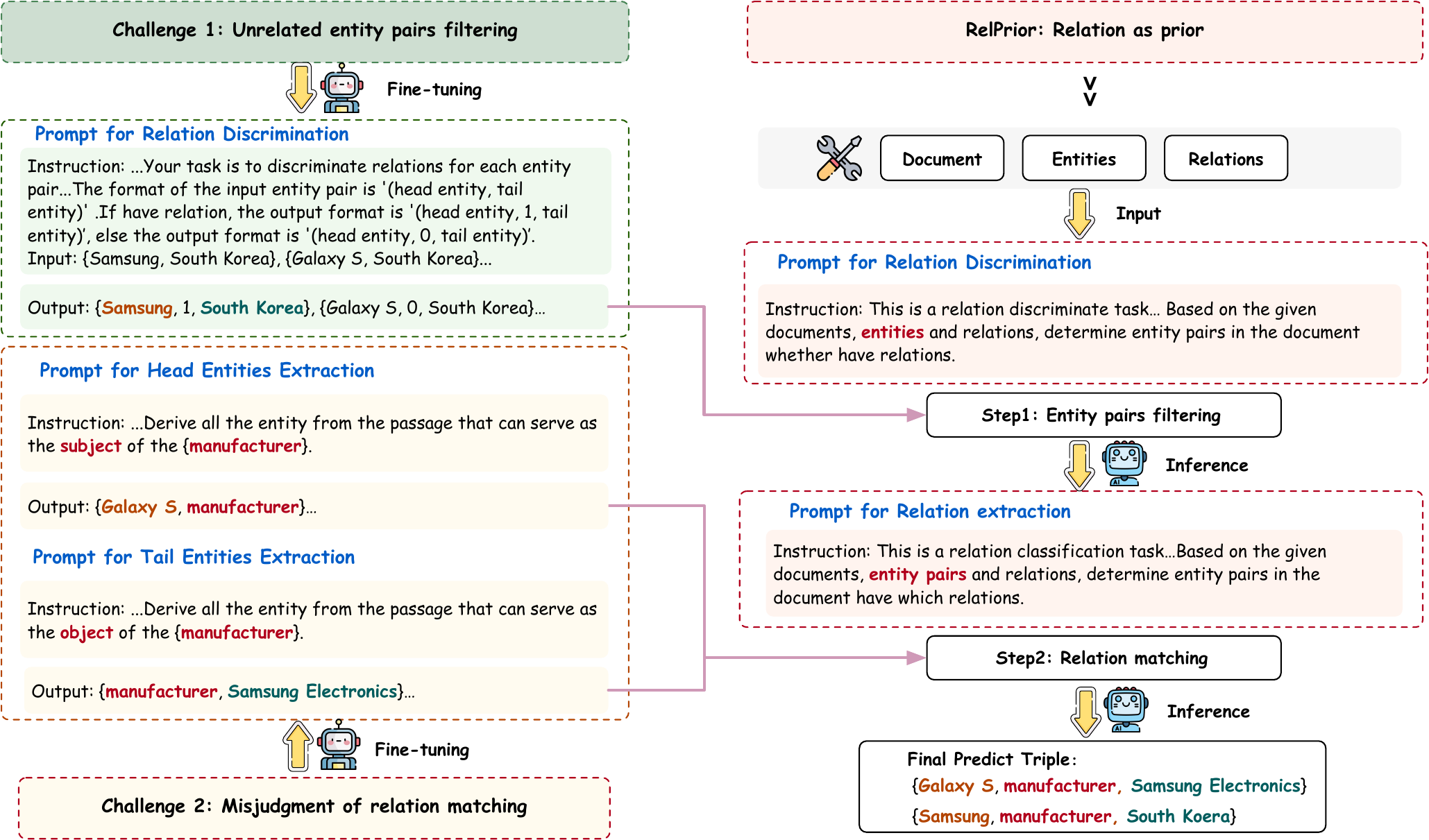}
  \caption {Our proposed Relation as a Prior (RelPrior) paradigm framework.}
  \label{Graph2}
\end{figure*}

\section{Methodology}

To address the challenges of applying LLMs in DocRE, we propose a novel relation as a prior paradigm for DocRE. As shown in Fig. \ref{Graph2}, the RelPrior architecture consists of two main modules: the first module filters out irrelevant entity pairs, the second module matches entity pairs based on the relations, as detailed below.

\subsection{Entity Pairs Filtering}
To mitigate the hallucination problem caused by unrelated entity pairs during LLMs reasoning, we propose the Entity Pairs Filtering (EPF) module. This module leverages the robust semantic reasoning capabilities of LLMs, capitalizing on their strong performance in few-shot classification tasks to determine whether relationships exist between entity pairs via binary classification.

Existing studies \cite{li2024llm} have shown that in order to enable LLMs to better adapt to specific tasks, fine-tuning LLMs is an indispensable step in handling DocRE tasks. By converting the format of supervised data into a question-answer form for determining the presence or absence of relationships between entity pairs, and inputting it into the model to learn the corresponding questions and answers, and conducting training in combination with context information. This processing method can not only enhance the model's understanding of the task objective, but also improve the model's focus on the relation presence or absence judgment task, thereby more effectively exerting the ability of the LLMs in classification tasks.

Furthermore, this module introduces a strategy of incorporating both positive and negative samples during fine-tuning.  Specifically, we extract not only positive samples (entity pairs with relationships) from the labeled data, but also negative samples (entity pairs without relationships) through random sampling.  By jointly training on these positive and negative samples, the model can better learn the distinguishing features between related and unrelated entity pairs, thereby improving classification accuracy. Specifically, the prompt content is as follows:

\begin{mybox}
	\texttt{\textbf{Instruction:} This is a relation discrimination task...If the relationship between the head entity and tail entity exists in the relation set, your output format is \textbf{‘(head entity, 1, tail entity)’}. \TheText{[The Text $\mathcal{L}_{text}$]}
    \RelationSet{[Relation set $\mathcal{L}_{re}$]}\\
        \textbf{Input:} \EntitySet{[Entity set $\mathcal{L}_{entities}$]}}
\end{mybox}

Specifically, for given \TheText{[The Text $\mathcal{L}_{text}$]}, \RelationSet{[Relation set $\mathcal{L}_{re}$]} and \EntitySet{[Entity set} \EntitySet{$\mathcal{L}_{entities}$]}, the Entity Pairs Filtering module guides LLMs to conduct semantic analysis and context understanding for each set of entity pairs. Then determine whether there is a relationship between them and output the set of entity pairs that have a relationship. In this way, the Entity Pair Filtering module can effectively filter out a large number of irrelevant entity pairs, reduce the input scale of subsequent relation discrimination tasks, and thereby lower the risk of hallucination problems in large models when processing irrelevant entity pairs information. The task of discriminating whether there is a relationship can be formally expressed as:
\begin{eqnarray}
\mathcal{L}_r^{\prime}=\underset{\mathcal{L}}{\operatorname*{\operatorname*{argmax}}}\sum_{i=1}^NP(\mathcal{L}_r^i\mid\mathcal{L}_{text},\mathcal{L}_{re},\mathcal{L}_{entities}^i)
\end{eqnarray}
where $\mathcal{L}_r^{\prime}=\{\mathcal{L}_r^{i}\}_{i=1}^N$ denotes eventually determine the entity pair that exists the relationship, $\mathrm{\textit{N}}$ denotes number of relationships present in the dataset, $\mathcal{L}_{entities}^i\in\mathcal{L}_{entities}$ denotes entity pairs that need to be judged for the existence of relations. $\mathcal{L}_{text}$ and $\mathcal{L}_{re}$ denote the input source document and the relation sets, respectively.

In addition, to better verify whether the method of directly extracting relations from the filtered entity pairs is feasible, we have also added a process of directly discriminating relations from the filtered entity pairs. This process aims to accurately identify relationships between the related entity pairs, establishing a solid foundation for subsequent triple construction. Designing specifically for RE tasks, this process reformats supervised data into a question-answer format better suited for LLMs processing while incorporating contextual information during training, significantly enhancing its relation classification performance. The detailed prompt structure is as follows:

\begin{mybox}
	\texttt{\textbf{Instruction:} This is a relation classification task...The format of the input entity pair is ‘(head entity, tail entity)’. Your output format is \textbf{‘(head entity, relation, tail entity)’}. \TheText{[The Text $\mathcal{L}_{text}$]}
    \RelationSet{[Relation set $\mathcal{L}_{re}$]}\\
        \textbf{Input:} \EntityPairs{[Entity Pairs $\mathcal{L}_r^{^{\prime}}$]}}
\end{mybox}

Specifically, by analyzing for the given document \TheText{[The Text $\mathcal{L}_{text}$]}, \RelationSet{[Relation} \RelationSet{set $\mathcal{L}_{re}$]}, and \EntityPairs{[Entity set $\mathcal{L}_{entities}$]} for which the Entity Pairs Filtering module initially determines the relationship. Then, this module guides the LLMs to further determine what correlation information might exist between these entity pairs to ensure that the correct relationships are extracted as much as possible. It can be formally defined as:

\begin{eqnarray}
F^{\prime}=\underset{f}{\operatorname*{\operatorname*{argmax}}}P(f\mid\mathcal{L}_{text},\mathcal{L}_{re},\mathcal{L}_{r}^{\prime})
\end{eqnarray}
where $\mathcal{L}_{text}$, $\mathcal{L}_{re}$ and ${\mathcal{L}}_{r}^{\prime}$ denote the input source document, the relation and the set of entity pairs obtained in previous steps for which the relation exists, respectively. $F^{\prime}$ denotes the predicted triple sets.

\subsection{Relation Matching}
To address the strict relation labels results statistical may be misjudged and the potential omission of valid triples in the model's initial predictions after filtering unrelated entity pairs, we construct the Relation Matching (RM) module. This module aims to use relation as a prior and discover additional plausible triplet facts based on the predicted relations, thereby improving the completeness of RE.

Specifically, firstly, construct the fine-tuning datasets of the head entity and the tail entity respectively to better meet the requirements of the head and tail entity inference tasks. Then, extract all the relations \PredictRelationSet{[Predict Relation set $r$]} from the triplet sets predicted in Section 3.1, and combine the given \TheText{[The Text} \TheText{$\mathcal{L}_{text}$]} and \Entity{[Entity $\mathcal{L}_{e}$]} as the model input. Based on the constraints of context semantics and relations, the model inferences the possible candidate sets of head entities ($h_{r}$, $r_{1}$) and tail entities ($r_{2}$, $t_{r}$) respectively. The head entity prompt construction is as follows:

\begin{mybox}
	\texttt{\textbf{Instruction:} Given \PredictRelationSet{[Predict Relation set $r$]}. Now the passage is \TheText{[The Text $\mathcal{L}_{text}$]}. Derive all the entity from the passage that can serve as the \textcolor{red}{subject} of the relation. Your output format is \textbf{{(entity1, relation1)…}}\\
        \textbf{Input:} \Entity{[Entity $\mathcal{L}_{e}$]}}
\end{mybox}

Similarly, the tail entity prompt construction is as follows:
\begin{mybox}
	\texttt{\textbf{Instruction:} Given \PredictRelationSet{[Predict Relation set $r$]}. Now the passage is \TheText{[The Text $\mathcal{L}_{text}$]}. Derive all the entity from the passage that can serve as the \textcolor{red}{object} of the relation. Your output format is \textbf{{(relation1, entity1)…}}\\
        \textbf{Input:} \Entity{[Entity $\mathcal{L}_{e}$]}}
\end{mybox}

After obtaining the corresponding candidate sets ($h_r$, $r_1$) and ($r_2$, $t_r$) based on the above analysis. The sets with the same relationship in ($h_r$, $r_1$) and ($r_2$, $t_r$) are merged to screen out the logical new triples ($h_r^{\prime}$, $r$, $t_r^{\prime}$). The formula for the triplet result $R^{\prime}$ is as follows:
\begin{eqnarray}
R^{\prime}=\{(h_r^{\prime}, r, t_r^{\prime})\mid h_r^{\prime}\in h_r,t_r^{\prime}\in t_r,r_1=r_2\}
\end{eqnarray}

To address potential redundancies and inconsistencies between the triples predicted by entity pairs filtering and head-tail entities inference, we integrate and refine their outputs, ultimately generating the final triplet set. This operates in two stages: First, it performs filtering on the outputs from both entity pairs filtering and head-tail entities inference, removing both duplicate triples and invalid triples containing relations absent from the predefined set. Second, during the fusion process, the triples from the entity pairs filtering module are prioritized due to their higher accuracy in relation identification, while high-quality supplementary triples from the relation matching module are incorporated to ensure broader coverage of relation facts.

\section{Experiments}
\subsection{Datasets}
To further verify the effectiveness of the method, we conducted evaluations on the document-level relation extraction public datasets DocRED\footnote{https://github.com/thunlp/DocRED.} \cite{yao2019docred} and RE-DocRED\footnote{https://github.com/tonytan48/Re-DocRED.} \cite{tan2022revisiting}. The specific statistical information such as the number of documents, entities, sentences, and triples of the two datasets are shown in Table \ref{table1}. Since the test set labels of the DocRED dataset are not public, the relevant information of the test set is not reported here.

The DocRED dataset covers 96 common types of relationships from Wikipedia, and the relationship facts are not restricted by any specific field, demonstrating strong universality. The RE-DocRED dataset relabels different mentions of entities based on the DocRED dataset, aiming to alleviate the problem of false negative examples that identify correct samples as negative samples existing in DocRED.

\begin{table}[ht]  
    \centering
    \vspace{-10pt}
    \caption{Statistics on the relevant information for the training set, development set, and test set of both DocRED and RE-DocRED.}
    \label{tab:dataset_stats}
    \setlength{\tabcolsep}{10pt} 
    \renewcommand{\arraystretch}{1.1} 
    \begin{tabular}{lccccc}
    \hline
    \multirow{2}{*}{Dataset} & \multicolumn{2}{c}{\textbf{DocRED}} & \multicolumn{3}{c}{\textbf{RE-DocRED}} \\
    \cmidrule(lr){2-3} \cmidrule(lr){4-6}
     & Train & Dev & Train & Dev & Test \\
    \hline
    Number of documents & 3053 & 1000 & 3053 & 500 & 500 \\
    Avg.\ entities per document & 19.5 & 19.6 & 19.4 & 19.4 & 19.6 \\
    Avg.\ sentences per document & 7.9 & 8.1 & 7.9 & 8.2 & 7.9 \\
    Avg.\ triples per document & 12.5 & 12.3 & 28.1 & 34.6 & 34.9 \\
    \hline
    \end{tabular}
    \label{table1}
    \vspace{-15pt}
\end{table}

\subsection{Baselines}

We adopt two categories of baseline models for evaluation: 

\textbf{Pre-trained Language Models(PLM)} use traditional small models to represent text as sequences or graphs for DocRE tasks, with BERT being the primary model.
\cite{zeng2020double} constructs two graphs, where the mention graph captures the complex interactions between different mentions, and the entity graph aggregates different mentions of the same entity. \cite{zhou2021document} proposes adaptive threshold and local context pool techniques to obtain the relevant context of entity pairs. \cite{xie2022eider} efficiently combines evidence sentences to enhance DocRE. \cite{ma2023dreeam} uses evidence sentence information as supervision signal to perform DocRE.

\textbf{Large Language Models(LLM)} use powerful language modeling and representation learning capabilities and conduct DocRE through natural language interaction. \cite{sun2024consistency} directly infers the possible relationships between related entity pairs. \cite{li2023semi} combines natural language reasoning methods for DocRE. \cite{sun2024consistency} uses the search chain prompt to extract the triplet relation. \cite{xue2024autore} uses the paradigm of document-relation-head-triple fact to extract relations. \cite{li2024llm} fine-tuning llama3-8b model to extract the corresponding triples directly from the document.

\subsection{Metric and experiment settings}
Consistent with other SOTA methods, we select F1 and Ign F1 as the main evaluation metrics. F1 takes into account both precision and recall comprehensively to evaluate the performance of the model more comprehensively. To avoid the model memorizing the relationship facts in the training set during the training process, we use Ign F1 for evaluation. This metrics excludes the existing relationship facts in the training data and more objectively reflects the actual performance of the model in the RE task.

We select five representative open-source pre-trained LLMs as basic models: DeepSeek-R1-Distill-LLaMA3-8B \cite{guo2025deepseek}, Qwen2.5-7B \cite{bai2023qwen}, LLaMA2-13B-Chat \cite{touvron2023llama}, LLaMA3.1-8B \cite{grattafiori2024llama} and LLaMA3-8B \cite{grattafiori2024llama}. All five models are implemented based on the LLaMA-Factory \cite{zheng2024llamafactory} architecture. In all steps, the top-p and temperature parameters are selected within the range of (0,1). All experiments are conducted on a Tesla V100s GPU with a single card of 32G video memory.

\subsection{Main Results}

\begin{table}[ht]
    \caption{Experimental results on the development and test set of DocRED and RE-DocRED datasets. Results with * are our local experiments using LLAMA3-8B model. Bold indicates the best results among the LLM-based methods.}
    \centering
    \begin{tabular}{lcccccccc}
    \toprule
    & \multicolumn{4}{c}{\textbf{DocRED}} & \multicolumn{4}{c}{\textbf{Re-DocRED}}\\
    \cmidrule(lr){2-5} \cmidrule(lr){6-9}
    \multirow{2}{*}{\textbf{Method}} &
    \multicolumn{2}{c}{\textbf{Dev}} &
    \multicolumn{2}{c}{\textbf{Test}} &
    \multicolumn{2}{c}{\textbf{Dev}} &
    \multicolumn{2}{c}{\textbf{Test}}\\
    \cmidrule(lr){2-3} \cmidrule(lr){4-5} \cmidrule(lr){6-7} \cmidrule(lr){8-9}
    & Ign $F_1$ & $F_1$ & Ign $F_1$ & $F_1$ & Ign $F_1$ & $F_1$ & Ign $F_1$ & $F_1$\\
    \midrule
        \textbf{PLM-based methods} \\
        GAIN\_Bert-base & 59.14 & 61.22 & 59.00 & 61.24 & - & - & - & - \\
        ATLOP\_Bert-base & 59.22 & 61.09 & 59.31 & 61.30 & 76.88 & 77.63 & 76.94 & 77.73 \\
        EIDER\_Bert-base & 60.51 & 62.48 & 60.42 & 62.47 & - & - & - & - \\
        DREEAM\_Bert-base & 59.60 & 61.42 & 59.12 & 61.13 & - & - & 77.34 & 77.94 
        \\
    \midrule
        \textbf{LLM-based methods} \\
        ChatGPT & - & 21.9 & - & 23.6 & - & 20.6 & - & 21.7\\
        DocGNRE\_LLaMA3-8B*& 11.22 & 11.60 & 12.69 & 12.35 & 11.86 & 9.53 & 11.03 & 9.39 \\
        CoR\_LLaMA3-8B*   & - & 39.02  & - & 38.45 & - & 40.85 & - & 39.74\\
        D-R-F\_LLaMA3-8B*& 40.02 & 41.66 & 40.36 & 41.97 & 50.56 & 51.04 & 50.02 & 50.49 \\
        AutoRE\_LLaMA3-8B*& 45.22 & 45.76 & 45.10 & 45.96 & 52.14 & 52.28  & 51.55 & 51.97 \\
        LoRA FT\_LLaMA3-8B* & 40.13 & 41.11 & 39.86 & 40.66 & 47.35 &47.66  & 44.85 & 45.04   \\
        \midrule
        \textbf{Our Methods (RelPrior)} \\
        DeepSeek-R1-Distill-LLaMA3-8B &38.82  &39.46 & 37.37 & 38.99 & 41.82 & 42.07 & 41.06 & 41.27\\
        Qwen2.5-7B &43.05  &44.38 & 41.22 & 43.93 & 51.02 & 51.43 & 50.64 & 50.72\\
        LLaMA2-13B-Chat   & 40.97 & 41.11 & 41.23 & 42.54 & 50.04 & 50.62 & 49.82 & 49.86\\
        LLaMA3.1-8B  & 45.97 & 46.03 & 44.77 & 47.42 & 54.31 & 54.88 & 53.48 & 53.56\\
        LLaMA3-8B    & \textbf{47.70} & \textbf{47.82} & \textbf{46.41} & \textbf{48.34} & \textbf{55.50} & \textbf{55.94} & \textbf{54.53} & \textbf{54.80}\\
    \bottomrule
    \end{tabular}
    \label{table2}
    \vspace{-10pt}
\end{table}%

Table \ref{table2} presents the overall performance comparison of each baseline model, the RelPrior architecture proposed in this paper on the DocRED and RE-DocRED datasets. It can be seen from the experimental results that the RelPrior\_LLaMA3-8B model adopting the RelPrior architecture significantly outperforms the existing methods based only on LLMs in the DocRE task, achieving the state-of-the-art performance.

Specifically, compared with the current method AutoRE\_LLaMA3-8B that only on LLMs, the F1 values of the RelPrior\_LLaMA3-8B model in the development sets and test sets of DocRED have increased by 2.06\% and 2.38\% respectively. Moreover, in the RE-DocRED dataset, RelPrior\_LLaMA3-8B has also achieved the optimal effect compared with the DocRE method based only on LLMs. This significant performance improvement demonstrates that RelPrior can effectively enhance the RE capabilities of LLMs. Moreover, the RelPrior paradigm can be readily adapted to different task requirements by either replacing the base LLMs or adjusting corresponding module parameters, showing strong potential for various DocRE applications.

It is worth noting that methods using only LLMs still lag behind those based on traditional pre-trained models like BERT, particularly in performance. This gap is primarily due to the limitations of current LLMs in handling a vast number of classification tasks, as well as the issue of hallucination. These challenges contribute to the poor performance of LLMs in document-level relation extraction. However, comparisons between different versions of LLMs show continuous performance improvements with each update. For example, the shift from LLaMA2 to LLaMA3 brought significant performance gains, suggesting that LLMs have great potential for document-level relation extraction tasks.

To verify the generalization of RelPrior across different open-source LLMs, we conducted experiments using RelPrior with two different types LLMs, Qwen2.5-7B and DeepSeek-R1-Distill-LLaMA3-8B. As shown in Table \ref{table2}, they achieved strong relation extraction performance on the DocRED and RE-DocRED test sets. The modular design of RelPrior allows it to adapt flexibly to different LLMs, effectively improving relation extraction performance across the board. This flexibility opens up the potential for applying RelPrior to more LLMs, with the prospect of better integration and performance in future models.

\begin{table*}[ht]
\caption{Ablation Study Results on the DocRED dataset. EPF represents Entity Pairs Filtering module, RM represents Relation Matching module.}
\centering
    \begin{tabular}{lllll}
    \toprule 
    \multirow{2}{*}{\bf Model} & \multicolumn{2}{c}{\bf Dev} & \multicolumn{2}{c}{\bf Test}  \\ 
    \cmidrule(lr){2-3} \cmidrule(lr){4-5}
    ~ & \bf Ign F1 &  \bf F1 & \bf Ign F1 & \bf F1 \\
    \midrule 
    RelPrior\_LLaMA3-8B  & \textbf{47.70} & \textbf{47.82} & \textbf{46.41}  &  \textbf{48.34} \\
    w/o (EPF)  &  27.73 \textcolor{red}{(-19.97)} &  28.84 \textcolor{red}{(-18.98)} &  26.32 \textcolor{red}{(-20.09)} & 26.98 \textcolor{red}{(-21.36)} \\
    w/o (RM)  & 42.91 \textcolor{red}{(-4.79)} &  43.02 \textcolor{red}{(-4.80)} &  41.37 \textcolor{red}{(-5.04)} &  43.67 \textcolor{red}{(-4.67)} \\
    \midrule 
    RelPrior\_LLaMA2-13B-Chat  &  \textbf{40.97} &  \textbf{41.11} &  \textbf{41.23} &  \textbf{42.54} \\
    w/o (EPF)   & 26.94 \textcolor{red}{(-14.03)} &  27.24 \textcolor{red}{(-13.87)} &  25.08 \textcolor{red}{(-16.15)} &  25.33 \textcolor{red}{(-17.21)}\\ 
    w/o (RM)  & 38.05 \textcolor{red}{(-2.92)} &  38.16 \textcolor{red}{(-2.95)}  &  37.88 \textcolor{red}{(-3.35)} &  38.99 \textcolor{red}{(-3.55)}\\ 
    \midrule 
    RelPrior\_LLaMA3.1-8B & \textbf{45.97} &  \textbf{46.03} &  \textbf{44.77}   &  \textbf{47.42}     \\
    w/o (EPF)  & 28.08 \textcolor{red}{(-17.89)} &  28.36 \textcolor{red}{(-17.67)}  &  27.22 \textcolor{red}{(-17.55)} &  28.13 \textcolor{red}{(-19.29)}\\ 
    w/o (RM)  & 43.12 \textcolor{red}{(-2.85)} &  43.26 \textcolor{red}{(-2.97)}  &  40.91 \textcolor{red}{(-3.86)} &  43.13 \textcolor{red}{(-4.29)}\\ 
\bottomrule
\end{tabular}
\label{table3}
\end{table*}

\subsection{Analysis}

To evaluate the effectiveness of RelPrior, we explored the following research issues:

\begin{itemize}
    \item[$\bullet$] RQ 1: Are all modules effective for the RelPrior architecture?
    \item[$\bullet$] RQ 2: Entity Pairs Filtering module have already obtained the predicted triples. Why proceed with the subsequent steps?
    \item[$\bullet$] RQ 3: How well does RelPrior perform on extracting different types of relations?
\end{itemize}

\subsubsection{Answer1: All modules are effective for the RelPrior architecture.}
This section conducts an in-depth analysis of the effectiveness of each module.

To verify the performance of the EPF module, a systematic ablation experiment was conducted on this module in this section. It can be seen from the w/o EPF results in Table \ref{table3} that when the EPF module is removed and directly used for RM operation, The performance of the RelPrior\_LLaMA3-8B model decreased by 18.98\% and 21.36\% respectively on the development sets and test sets of the DocRED dataset. This significant performance decline indicates that the EPF module plays a crucial role in the RE task. Through detailed analysis, we found that the EPF module's filtering of irrelevant entity pairs allows the model to better focus on potentially meaningful entity pairs. This mechanism effectively prevents the model's attention from being dispersed by irrelevant pairs while enhancing its discriminative capability for key entity pairs. In contrast, the EPF module first filters out irrelevant pairs before performing relation discrimination, significantly reducing task complexity. This progressive approach enables the model to gradually refine its task objectives, leading to more efficient RE.

To verify the functionality and effectiveness of the RM combination, it can be seen from the w/o RM results in Table \ref{table3} that when the RM module is removed, it can be observed that the Ign F1/F1 values of the development sets and the test sets decrease by 4.79\%/4.80\% and 5.04\%/4.67\% respectively. Through analysis, it is believed that only using EPF module may lead to incomplete triple extraction of certain specific relationships, especially when there are indirect implicit relationships in the document. The RM module can mine more possible triplet facts, while reducing the computational cost of the model and enabling it to focus more on potential RE tasks. Furthermore, the RM module effectively eliminates duplicate and conflicting triplets by filtering and fusing the triplet set, thereby enhancing the overall performance of the model.

\subsubsection{Answer2: The filtering of irrelevant entity pairs may lead to the absence of triplet facts.}

\begin{figure}[t]
  \centering
  \includegraphics[width=0.7\textwidth]{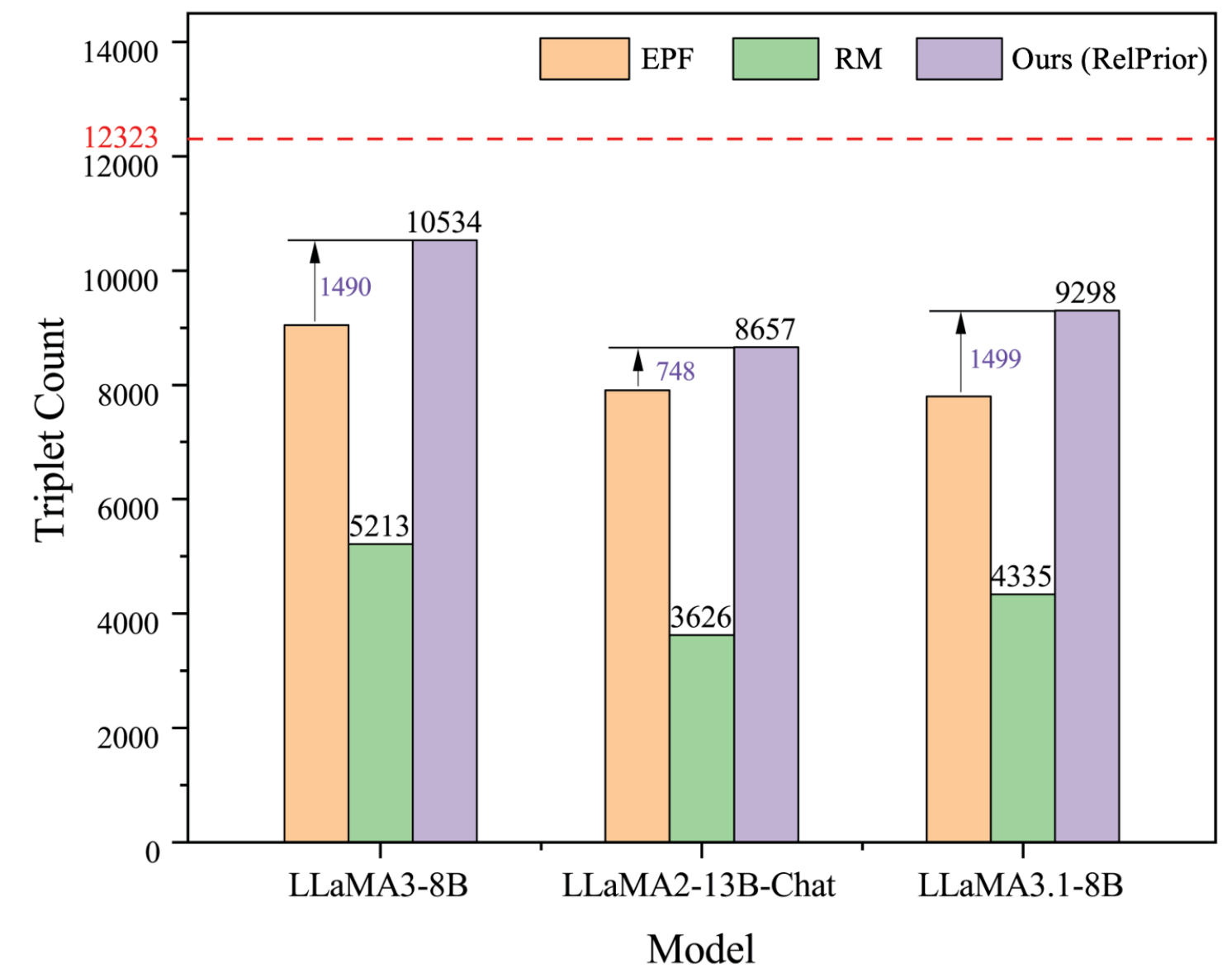}
  \caption{The F1 scores of each relationship type in the DocRED development sets.}
  \label{Graph6}
\end{figure}

Although the EPF module significantly improves extraction accuracy, it may miss many valid triples. 

To ensure the completeness of relation extraction, we introduce the RM module. Fig. \ref{Graph6} illustrates the number of extracted triples at various processing stages across different models on the DocRED development set. Taking LLaMA3-8B as an example, the EPF module initially identifies 9,044 triples. Subsequently, the RM module predicts potential head-tail entity pairs based on these identified relationships, generating an additional 5,213 triples. After filtering and merging through the TM module, the system produces a final set of 10,534 triples. This result is significantly closer to the 12,323 annotated triples in the development set, demonstrating the effectiveness of the model in extracting a more complete set of relations. Moreover, the approach maintains a notable improvement in relation extraction accuracy, as it captures more nuanced relationships that might otherwise have been overlooked, further enhancing the overall performance of the system.

\begin{figure}[t]
  \centering
  \includegraphics[width=0.7\textwidth]{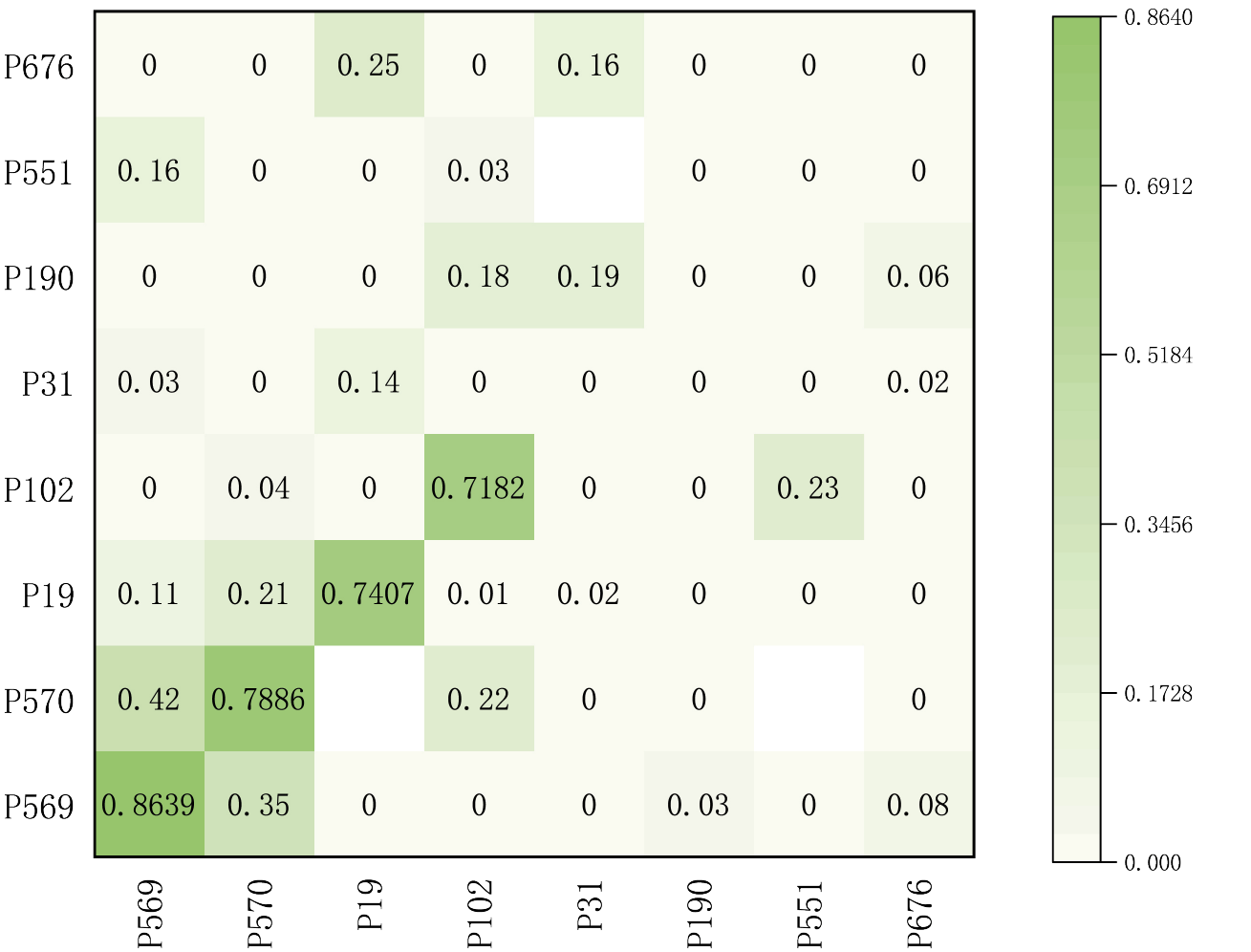}
  \caption{The F1 scores of relation type in the DocRED development sets.}
  \label{Graph5}
\end{figure}

\subsubsection{Answer3: There are differences in RelPrior in different relationship predictions.}
The extraction effects of the RelPrior on different relation vary to some extent.

To deeply analyze the distribution characteristics of relation, this paper is based on the prediction results of RelPrior\_LLaMA3-8B. As shown in Fig. \ref{Graph5}, four relationships with the best performance were selected: \textit{P569} (\textit{"date of birth"}), \textit{P570} (\textit{"date of death"}), \textit{P19} (\textit{"place of birth"}), and \textit{P102} (\textit{"member of political party"}). Fine-grained analysis was conducted on the four worst-performing relationships, namely \textit{P31} (\textit{"instance of"}), \textit{P190} (\textit{"sister city"}), \textit{P551} (\textit{"residence"}), and \textit{P676} (\textit{"lyrics by"}). 

For the relationships with excellent performance, this is mainly because such relationships are widely distributed in the training data, and the relationship expressions have clear statistical characteristics, which is convenient for the model to capture. The poor performance of certain relationships can be primarily attributed to their infrequent occurrence in the training corpus, as well as the relatively implicit nature of their expressions, which makes them prone to misjudgment. As a result, the model struggles to correctly generalize these relationships. Therefore, the generalization performance of the current model on low-frequency relation types still requires significant improvement, particularly in terms of its ability to handle nuanced or less explicit relational expressions.

\begin{figure}[t]
  \centering
  \includegraphics[width=1\textwidth]{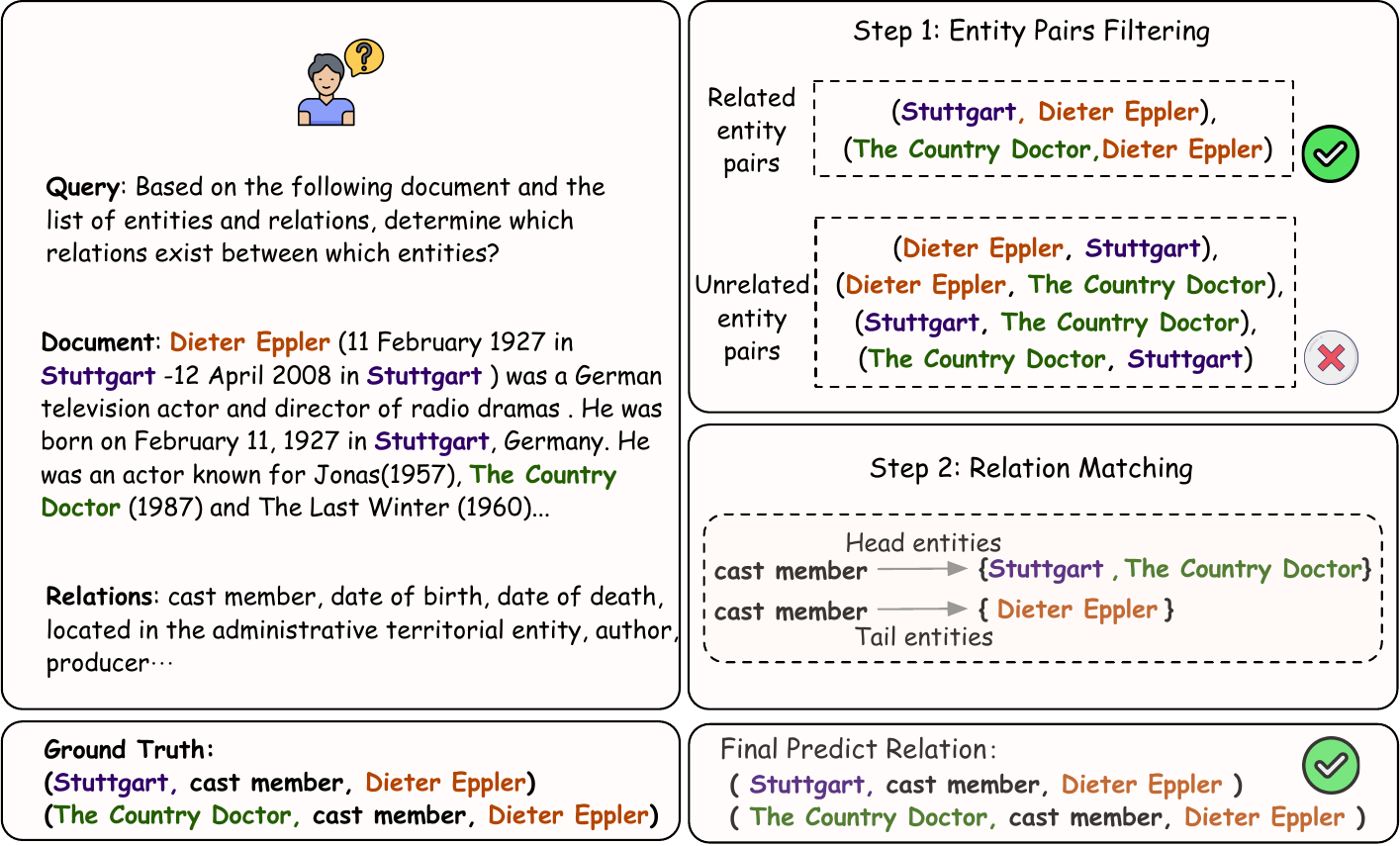}
  \caption{The case study of our proposed RelPrior architecture.}
  \label{Graph7}
\end{figure}

\subsection{Case Study}
To understand the effect of the RelPrior architecture more intuitively, this section conducts verification and analysis through a specific case in the DocRED dataset. As shown in Fig. \ref{Graph7}, when the RelPrior is adopted, the model removes the unrelated entity pairs \textit{\{Dieter Eppler, Stuttgart\}}, \textit{\{Dieter Eppler, The Country Doctor\}}, \textit{\{Stuttgart, The Country Doctor\}} and \textit{\{The Country Doctor, Stuttgart\}}. Then, based on relation matching module, the head entities associated with the relation (\textit{"cast member"}) are identified as \textit{Stuttgart} and \textit{The Country Doctor}, and the tail entities associated with this relation are \textit{Dieter Eppler}. Then the two are combined to obtain the final set of triples. This section shows that RelPrior not only effectively reduces the interference of irrelevant entities during model inference, but also successfully avoids misjudgments of triplets caused by strict relation labels, thereby significantly improving the model's overall relation extraction performance.

\section{Conclusion}
In this paper, we propose a novel document-level relation extraction paradigm, RelPrior, designed to enhance the performance of LLMs on this task. RelPrior consists of two components: Entity Pairs Filtering and Relation Matching. The first component filters out irrelevant entity pairs, allowing LLMs to focus on those with valid entity pais. This reduces complexity, enabling more accurate predictions. The second component leverages relationships between entity pairs to predict head and tail entities efficiently, uncovering additional triplet facts. This enhances the comprehensiveness of relation extraction, allowing the model to extract a broader range of relations with higher precision. Extensive experiments show that RelPrior outperforms existing LLM-based document-level relation extraction methods, proving its effectiveness in complex tasks.

\bibliographystyle{splncs04}
\bibliography{reference}
\end{document}